\documentclass[10pt,twocolumn,letterpaper]{article}

\usepackage{cvpr}
\usepackage{times}
\usepackage{array,multirow,graphicx}
\usepackage{amsmath}
\usepackage{amssymb}
\usepackage{booktabs}
\usepackage{siunitx}
\usepackage{caption}
\usepackage{subcaption}
\usepackage[export]{adjustbox}
\usepackage{float}
\usepackage{cuted}
\usepackage{capt-of}
\usepackage{lipsum}
\usepackage{dblfloatfix}
\usepackage{authblk}

\usepackage[breaklinks=true,bookmarks=false]{hyperref}

\cvprfinalcopy 

\newcommand*{\affaddr}[1]{#1} 
\newcommand*{\affmark}[1][*]{\textsuperscript{#1}}
\newcommand*{\email}[1]{\texttt{#1}}

\begin{document}

\title{Deep Fusion Network for Image Completion}

\author{%
Xin Hong\affmark[1]\thanks {This work is done when Xin Hong is an intern at Megvii Technology.}, \ Pengfei Xiong\affmark[2], \ Renhe Ji\affmark[2], \ and Haoqiang Fan\affmark[2]\\
\affaddr{\affmark[1]Institute of Computing Technology, Chinese Academy of Sciences} \quad
\affaddr{\affmark[2]Megvii Technology}\\
\email{hongxin@ict.ac.cn} \quad
\email{\{xiongpengfei, jirenhe, fhq\}@megvii.com}
}

\maketitle


\begin{strip}\centering
\includegraphics[width=\textwidth]{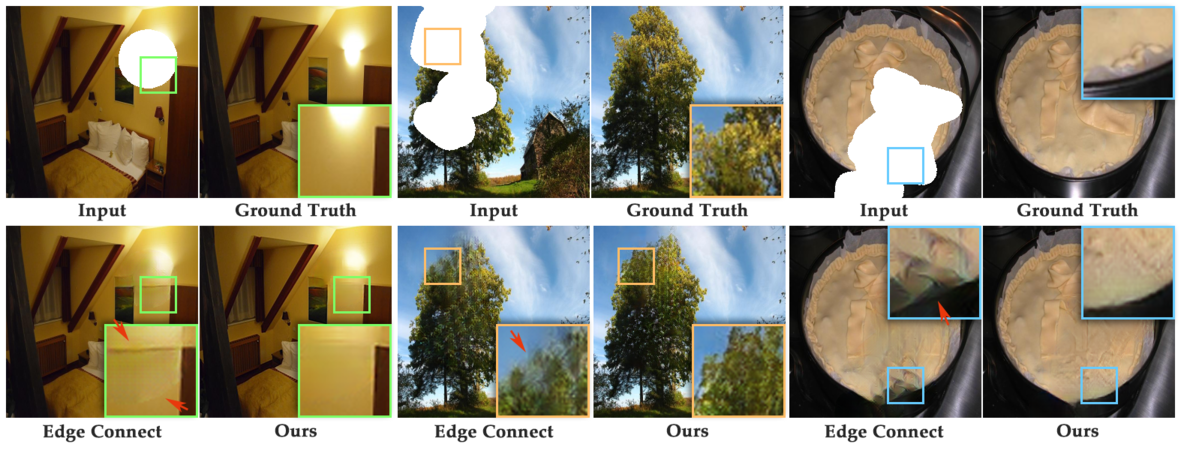}
\captionof{figure}{Comparison results between DFNet and previous state-of-the-art method Edge Connect\cite{nazeri2019edgeconnect}. In the first image of each group, white pixels represent the unknown region. With fusion blocks along with multi-scale constraints, DFNet has smoother transition (1st case), more natural texture (2nd case) and more consistent structure (3rd case).
\label{fig:teaser}}
\end{strip}

\begin{abstract}

Deep image completion usually fails to harmonically blend the restored image into existing content, especially in the boundary area. This paper handles with this problem from a new perspective of creating a smooth transition and proposes a concise Deep Fusion Network (DFNet). Firstly, a fusion block is introduced to generate a flexible alpha composition map for combining known and unknown regions. The fusion block not only provides a smooth fusion between restored and existing content, but also provides an attention map to make network focus more on the unknown pixels. In this way, it builds a bridge for structural and texture information, so that information can be naturally propagated from known region into completion. Furthermore, fusion blocks are embedded into several decoder layers of the network. Accompanied by the adjustable loss constraints on each layer, more accurate structure information are achieved. We qualitatively and quantitatively compare our method with other state-of-the-art methods on Places2 and CelebA datasets. The results show the superior performance of DFNet, especially in the aspects of harmonious texture transition, texture detail and semantic structural consistency. Our source code will be avaiable at: \url{https://github.com/hughplay/DFNet}
\end{abstract}

\section{Introduction}

Image completion, which aims to fill unknown region of an image, is a fundamental task in computer vision. It can be broadly applied to the fields of image editing, such as old photo recovering, object removal, and seamless inpainting for damaged image. For most such applications, it is a critical problem to generate perceptually plausible completion results, specifically with natural transition between known and unknown region.

Previous approaches based on deep learning have shown great progress in image completion task \cite{liu2018partialinpainting, yu2018generative, yu2018free, IizukaSIGGRAPH2017, nazeri2019edgeconnect, pathakCVPR16context, song2018spg, Yuhang2018Contextualinpaint, yiwang2018MultiColumn, Huy2018Structural, Haoran2018Progressive, Raymond2018Semantic}. As mentioned in \cite{bertalmio2000image}, these methods can be divided into two groups. One group of works focus on building a contextual attention architecture or applying effective loss functions to generate more realistic content in the missing area. They assume the gaps should be filled with similar content from background. A typical arrangement is applying Partial Convolutions\cite{liu2018partialinpainting} to concentrate on the unknown region. Other methods regard structural consistency as more important thing. Context priors such as edges are the most frequently used in these methods to ensure structural continuity. For instance, \cite{nazeri2019edgeconnect} proposed the Edge Connect method which can recover images with good semantic structural consistency. These approaches is dedicated to infer the unknown region with visually realistic and semantically related content. However, realizing smooth transition is more critical than restoring texture-rich images in most scenarios, as shown in Figure \ref{fig:teaser}.

Humans has an incredible ability to detect discontinuous transition region. Consequently, The filled region must be perceptually plausible in the transition zone with sufficiently similar texture and consistent structure. In order to achieve smooth transition, \cite{perez2003poisson} proposed a method to iteratively optimize the pixel gradient in edge transitional region. Given two images, the fusion quality depends on the consistency of the gradient changes of these two images, which is similar with the relationship between the restored content and the known region in image completion. This inspires us to build a network to simulate the composition process.

\begin{figure}
	\centering
	\includegraphics[width=1\linewidth]{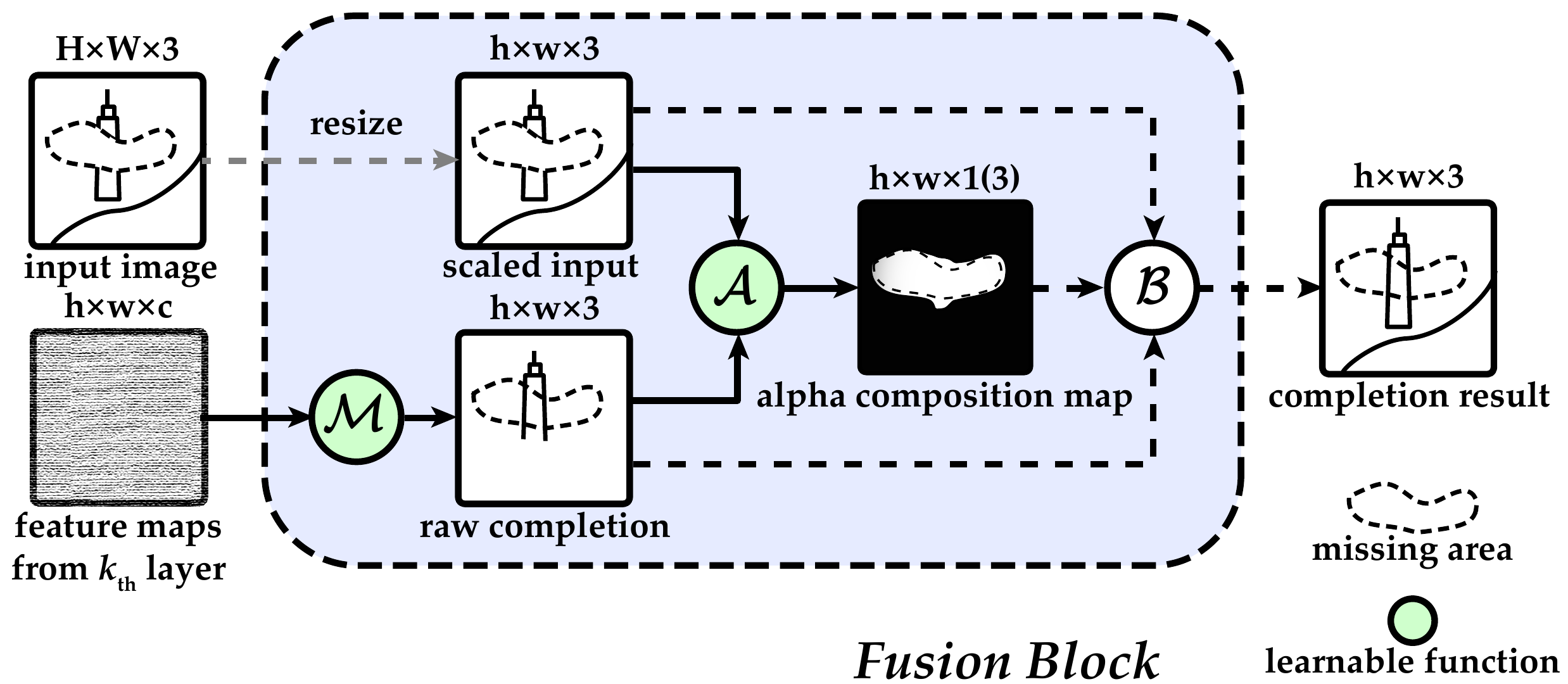}
	\caption{Illustration of Fusion Block.
    A fusion block extracts raw completion from feature maps by learnable function $\mathcal{M}$, and also predicts an alpha composition map with  function $\mathcal{A}$. Finally it combines the raw completion
    with scaled input image by blending function $\mathcal{B}$.
    The detail of blocks can be found in Section \ref{sec:fusion_block}.}
    \label{fig:fusion-block}
\end{figure}

In this work, we design a learnable fusion block to implement pixel level fusion in the transition region. As shown in Figure \ref{fig:fusion-block}, the fusion block is introduced that can be embedded to an encoder-decoder structure. Different from the previous methods, we develop an extra convolutional block to generate an alpha map, which is similar to the hole mask but has smoother weights especially on the boundary region. In the process of gradient descent optimization, the alpha composition map adjusts the balance between restored image and ground truth content to make the transition smoother. Similar ideas have also been used in image matting \cite{ImageMatting, NingXu_DIM}. However, The purpose of these method is to extract the smooth coefficients from background and foreground images, while the proposed fusion block is to combine them together. 

\begin{figure*}[h]
	\begin{center}
		\includegraphics[width=0.98\linewidth]{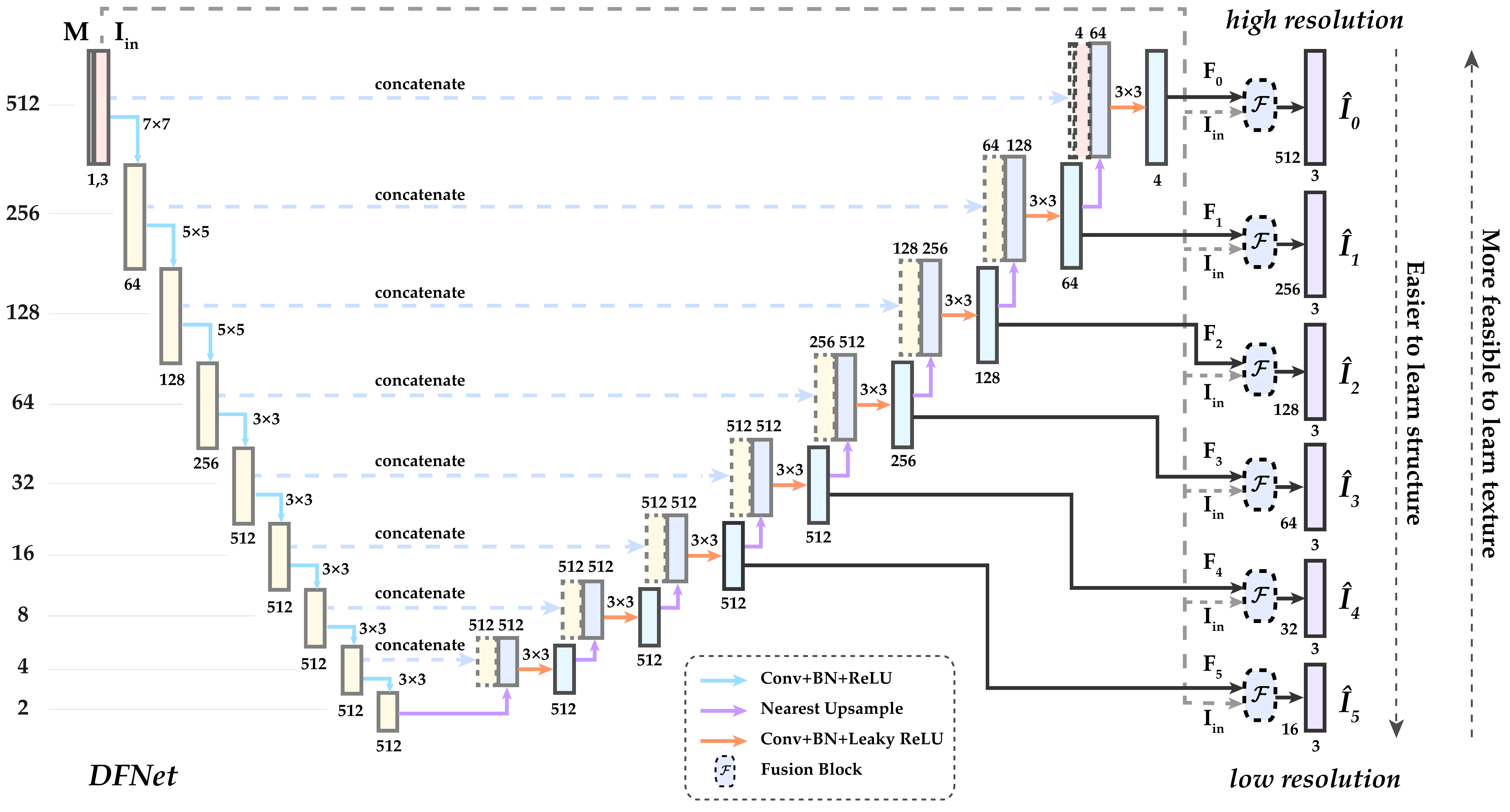}
	\end{center}
	\caption{Overview of our \textit{Deep Fusion Network (DFNet)}.
	DFNet is based on a U-Net, like the one used in \cite{pix2pix2017, liu2018partialinpainting}. The difference to traditional U-Net is that
	we embed fusion blocks to the last few decoder layers.
	During training, each fusion block will produce a completion result $\hat{\mathbf{I}}_k$
	from corresponding feature maps,
	which also has the same resolution with the feature maps.
	So that different constraints can be provided to each completion result as needed.
	During testing, only the completion result from last layer need to be produced.}
	\label{fig:dfn}
\end{figure*}

In detail, we propose a Deep Fusion Network (DFNet). Firstly, a fusion block is adopted as an adaptable module to combine the restored part of image and original image. In addition to providing a smooth transition, the fusion block avoids learning unnecessary identity mapping for pixels in unknown region, and provides an attention map to make network focusing more on the missing pixels. With fusion block, structural and texture information can be naturally propagated from known region into unknown region. Secondly, we embed this module into different decoder layers. We find out that by considering the prediction of different fusion blocks with multi-scale constraints, the deep fusion network outperforms the network with only one fusion block embedded to the final layer. Furthermore, while different layers provides different feature presentations, we selectively switch on and off structure and texture loss, to recover the structural information from lower layers and refine texture details in high layers. The whole architecture of DFNet is displayed in Figure \ref{fig:dfn}. 

The proposed DFNet is evaluated on two standard benchmarks, Places2 and CelebA. 
In order to better verify the proposed method, we define Boundary Pixels Error to measure the transition performance near the boundary of unknown region. Also, $\ell_{1}$ and FID are applied to verify global texture and consistency. Experiments demonstrate the superior performance of DFNet while compared with other state-of-the-art methods both in quantitative and qualitative aspects. It achieves better results in not only smooth texture transition but also structural consistency and more detailed textures.
As conclusion, the main contributions can be summarized as follows:
\begin{itemize}
\item We investigate the image completion problem with the perspective of better transition region and propose fusion block which predicts an alpha composition map to achieve smooth transition.
\item Fusion block avoids learning unnecessary identity mapping for known region and provides an attention mechanism. In this way, structure and texture information can propagate from known region to completion more naturally.
\item We propose Deep Fusion Network, a U-Net architecture embedded with multiple fusion blocks to apply multi-scale constraints.
\item A new measurement Boundary Pixels Error (BPE) is introduced to measure the transition performance near the boundary of missing hole.
\item The results on Places2 and CelebA show that our method outperforms state-of-the-art methods in both qualitative and quantitative aspects.
\end{itemize}

\section{Related Work}

\textbf{Context Aware}
Context aware based image completion methods imagine the semantic content can be filled based on the overall scene. Context Encoders\cite{pathakCVPR16context} introduces a encoder-decoder network to restore images from damaged inputs and holes. It applies a discriminator to increase the authenticity of restored images. Yang et al.\cite{Yang_2017_CVPR} takes its result as input and then propagates the texture information from unknown region to fill the missing area. Li et al.\cite{GFC-CVPR-2017} and Iizuka et al.\cite{IizukaSIGGRAPH2017} extends Context Encoders by defining both global and local discriminators to pay more attention on the missing areas. Iizuka et al. applies Poisson Blending\cite{perez2003poisson} as post-processing. Liu et al. \cite{liu2018partialinpainting} introduces partial convolution layers to avoid capturing too many zeros from unknown region. These methods depend entirely on the training image to generate semantically relevant structures and texture confidence.

\textbf{Texture Generation}
In the field of texture generation, perceptual loss is adopted to fill in visually realistic content for missing regions. Liu et al.\cite{liu2018partialinpainting} applies perceptual loss\cite{ gatys2015neural, Johnson2016Perceptual} which uses a VGG\cite{simonyan2014very} network as a feature extractor. It computes loss use extracted high level features to achieve higher resolution textures in completion. Other methods usually rely on GAN\cite{goodfellow2014generative} loss to obtain better details. For instance, Yu et al.\cite{yu2018generative} replaces the post-processing with a refinement network powered by the contextual attention layers. 

\textbf{Structure constraints}
To better control the completing behaviour of networks, other works\cite{song2018spg,yu2018free,nazeri2019edgeconnect} explore providing extra information for inpainting.
Song et al.\cite{song2018spg} uses a DeepLabv3+\cite{chen2018encoder} model to first predict a segmentation map, and then completes the unknown region with predicting segmentation map as prior. Yu et al.\cite{yu2018free} proposes gated convolution which generalize partial convolution and the new structure is compatible with user guides, usually strokes to indicate edges. Like Song et al.\cite{nazeri2019edgeconnect} uses a two staged networks for completion. It first completes edges corresponding to the input image and then use completed strokes to guide the full color images. In some extent, those methods can manually control the completion result of network by replacing the priors with custom one or giving extra edge information.

\textbf{Image Embedding}
As a similar work with image completion, image embedding and matting are also studied in the past decades. \cite{perez2003poisson} proposes a method to iteratively optimize the pixel gradient in edge transitional region. Then poisson matting\cite{sun2004poisson} firstly introduces a Poisson blending method into alpha matting by solving a poisson equation, which proves the effectiveness of alpha composition. Deep Image matting\cite{NingXu_DIM} also generates an alpha map with a encoder-decoder network. Cho et al\cite{ImageMatting} takes the matting results of \cite{closed-form} and normalized RGB colors as inputs and learn an end-to-end deep network to predict a new alpha matte. These methods prove that alpha matting based on deep learning is more realistic for image embedding and matting.

\begin{figure}
	\centering
	\includegraphics[width=0.8\linewidth]{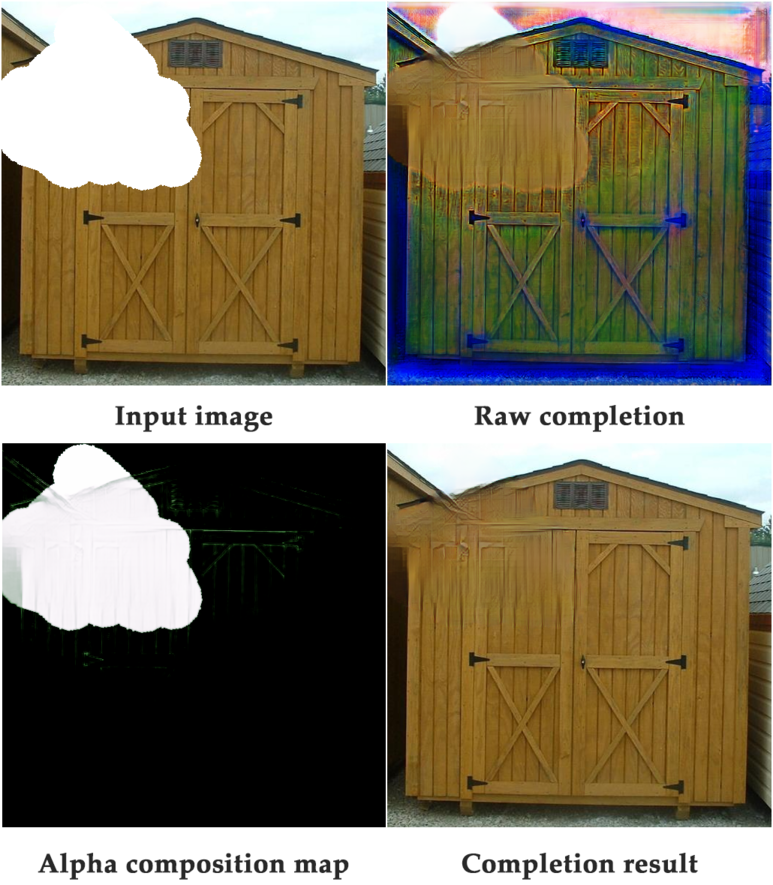}
	\caption{Corresponding results in a fusion block.}
	\label{fig:sample-alpha}
\end{figure}

\section{Deep Fusion Network}

\textit{Deep Fusion Network} is built on a U-Net\cite{ronneberger2015u} like architecture, which is widely used in recent image segmentation\cite{Long_2015_CVPR} task and image to image translation\cite{pix2pix2017, wang2018pix2pixHD, StarGAN2018, Kupyn_2018_CVPR} tasks. The difference between our DFNet and original U-Net is that we embed fusion blocks to several layers of decoder.
Fusion blocks help us to achieve smoother transition near the boundary and is the key components for our multi-scale constraints. In this section, we first introduce the fusion block and then discuss our network architecture and loss functions.
\subsection{Fusion Block}\label{sec:fusion_block}

The task of image completion is to restore the missing area
with visually plausible content from a damaged image $\mathbf{I}_{in}$ and a binary mask $\mathbf{M}$ which represents the location of the unknown region.

Recently deep learning based methods usually predict
the whole image $\mathbf{I}_{out}$ which even includes known region
and use it to calculate loss during training.
However, they take $\mathbf{I}_{comp}$ ($\mathbf{I}_{comp} = \mathbf{M} \odot \mathbf{I}_{in} +
(\mathbf{1} - \mathbf{M}) \odot \mathbf{I}_{out}$, $\odot$ denotes Hadamard product)
rather than $\mathbf{I}_{out}$ for testing.
The composition process replaces known region in $\mathbf{I}_{out}$
with corresponding pixels in $\mathbf{I}_{in}$.
Only a few methods\cite{liu2018partialinpainting} use both $\mathbf{I}_{out}$ and $\mathbf{I}_{comp}$ to compute loss.

This training strategy has problems.
Firstly, the mission of image completion is to complete the unknown region only. It is actually hard to complete missing hole while keeping a strict identity mapping for known area.
Secondly, the inconsistent use of $\mathbf{I}_{comp}$ and $\mathbf{I}_{out}$ during training and testing, along with the rigid composition method, usually produces visible artifacts around the boundary of missing area. As shown in the first case of Figure \ref{fig:teaser}, the result of Edge Connect\cite{nazeri2019edgeconnect} has a clear edge at the boundary of completion.

To remove the artifacts around the boundary and avoid neural networks learning unnecessary identity mapping, we propose \textit{Fusion Block}.
As shown in Figure \ref{fig:fusion-block}, a fusion block feed with two elements, an input image with unknown region $\mathbf{I}_{in}$
and feature maps $\mathbf{F}_k$ form $k_{\text{th}}$ layer
($1_\text{st}$ layer is the last decoder layer of U-Net).
The fusion block first extracts raw completion $\mathbf{C}_k$ from feature maps,
and then predicts an alpha composition map $\boldsymbol{\alpha}_k$ to combine them.
The final result $\mathbf{\hat{I}}_k$ is obtained by:
$$
\mathbf{\hat{I}}_k = \mathcal{B}(\boldsymbol{\alpha}_k, \mathbf{C}_k, \mathbf{I}_{k}) 
= \boldsymbol{\alpha}_k \odot \mathbf{\mathbf{C}_k}
+ (1 - \boldsymbol{\alpha}_k) \odot \mathbf{I}_{k}
$$
We resize $\mathbf{I}_{in}$ to obtain $\mathbf{I}_{k}$. The raw completion $\mathbf{C}_k$ extracted from feature maps $\mathbf{F}_k$
by a learnable function $\mathcal{M}$:
$$
\mathbf{C}_k = \mathcal{M}(\mathbf{F}_k)
$$
$\mathcal{M}(\mathbf{x})$ transforms $n$ channel feature maps $\mathbf{x}$
into a 3 channel image with the resolution unchanged
which is exactly the raw completion.
Actually, we use a $1\times 1$ convolutional layer following with a sigmoid function
to learn $\mathcal{M}$.

The alpha composition map $\boldsymbol\alpha_k$ is 
produced by another learnable function $\mathcal{A}$
from raw completion and the scaled input image:
$$
\boldsymbol{\alpha}_k = \mathcal{A}(\mathbf{F}_k, \mathbf{I}_{k})
$$
$\boldsymbol\alpha_k$ has two choices in the number of channels,
either single channel for image-wise alpha composition
or 3 channels for channel-wise alpha composition.
In practice, we find channel-wise alpha composition performs better.
As for $\mathcal{A}$, we use three convolutional layers
and the kernel size of them are 1, 3, 1.
First two convolutional layers are followed
with a Batch Normalization\cite{pmlr-v37-ioffe15} layer
and a leaky ReLU function.
And we apply sigmoid function to the output of last convolutional layer.

The fusion block enables network to avoid learning unnecessary identity mapping
while completing unknown region with soft transition near the boundary.
We also give an example of corresponding images in a fusion
block in Figure \ref{fig:sample-alpha}.
Completion performance can be further improved with multi-scale constraints by embedding fusion blocks to the last few decoder layers of U-Net.

\subsection{Network Architecture}

It's intuitive that when completing an image, 
constructing structures is easier in lower resolution for algorithms,
while recovering texture is more feasible in higher resolution.
We embed fusion blocks to the last few decoder layers of the U-Net 
and obtain completion results in different resolution.
And then we can apply structure and texture constraints to different resolution as we want.
The overview of our DFNet is shown in Figure \ref{fig:dfn}.
We choose U-Net\cite{ronneberger2015u} like the one used in
\cite{pix2pix2017, liu2018partialinpainting} as our backbone architecture.
The difference is that the last few decoder layers are embedded with fusion blocks.
Each fusion block outputs a completion result $\mathbf{C}_i$
with the same resolution as the input feature maps $\mathbf{F}_i$.
According to their resolution, we can provide different constraints as we want during training.
We will discuss these constraints in Section \ref{sec:loss}.
During testing, only the completion result $\mathbf{\hat{\mathbf{I}}_0}$ from last layer is needed.

\subsection{Loss Functions}\label{sec:loss}

The target of image completion is to generate visually plausible results
in both aspects of structure and texture.
Reconstruction loss, which is mean absolute error of each pixel
between prediction and ground truth,
is usually used to guarantee accurate structures in completion results.
However, high resolution textures is beyond the capability of reconstruction loss.
Previous works use GAN loss \cite{goodfellow2014generative}
or perceptual loss along with style loss \cite{Johnson2016Perceptual} to obtain vivid textures.
These two loss have same drawback which is known as
producing checkerboard and fish scale artifacts\cite{liu2018partialinpainting}.
Total variation loss is usually used to counter this drawback.
Results from \cite{liu2018partialinpainting} shows that this artifact
can be reduced more obviously by increasing the weight of style loss.

\textbf{Reconstruction Loss} Reconstruction loss is defined as mean absolute error
of completion result $\hat{\mathbf{I}}_k$ and target image $\mathbf{I}_k$:
$$
\mathcal{L}_{\ell_1}^k = \frac{1}{C_{k}H_{k}W_{k}} \lVert \mathbf{I}_k - \hat{\mathbf{I}}_k \rVert_1
$$
The number of channels is $C$, the height is $H$ and the width is $W$.


\textbf{Perceptual Loss and Style Loss}
Perceptual loss and style loss are first used
in style transfer\cite{gatys2015neural,Johnson2016Perceptual}.
They use a pre-trained VGG network to extract high level features.
The errors are computed between these features rather than original images.
Let $\phi_j(x)$ be the features of $j$th layer in a VGG network when given image $x$.
The size of $\phi_j(x)$ is $C_j\times H_j \times W_j$.
Perceptual loss is defined as the error of these features:
$$
\mathcal{L}_{p}^k = \sum_{j\in J}\lVert \phi_j(\mathbf{I}_k) - \phi_j(\hat{\mathbf{I}}_k) \rVert_1
$$
$J$ is selected VGG layers.
\textit{Gram matrix} is a $C_j \times C_j$ matrix, whose elements are defined as:
$$
G_{j}^{\phi}(x)_{c,c'} = \frac{1}{C_{j}H_{j}W_{j}}\sum_{h=1}^{H_j}\sum_{w=1}^{W_j}\phi_{j}(x)_{h,w,c}\phi_{j}(x)_{h,w,c'}
$$
And then style loss is $L_1$ mean absolute error between
corresponding Gram matrices of the output and target image:
$$
\mathcal{L}_{s}^k = \sum_{j\in J}\lVert G_{j}^{\phi}(\mathbf{I}_k) - G_{j}^{\phi}(\hat{\mathbf{I}}_k) \rVert_1
$$
Style loss doesn't consider the position of pixels but
cares about how high level features appear simultaneously\cite{Johnson2016Perceptual},
so that it's better for constraining the entire style of an image.

\textbf{Total Variation Loss} Total variation loss $\mathcal{L}_{tv}$
is errors computed only use predictions.
Each pixel will compute errors with top pixel and left pixel respectively.
This can be implemented more easily by using a convolution layer with a fixed kernel.

\textbf{Total Loss.} We group loss functions into \textit{Structure Loss} and \textit{Texture Loss}.
Structure Loss is represented as weighted reconstruction loss:
$$
\mathcal{L}_{struct}^{k} = \lambda_{\ell_1} \mathcal{L}_{\ell_1}^k
$$
And texture loss is a combination of three loss:
$$
\mathcal{L}_{text}^{k} = \lambda_{p} \mathcal{L}_{p}
+ \lambda_{s}\mathcal{L}_{s}^k + \lambda_{tv}\mathcal{L}_{tv}^k
$$
Our final loss is sum of structure loss and texture loss
from different resolution completion results:
$$
\mathcal{L}_{total} = 
\frac{1}{\lvert P \rvert}
\sum_{k\in{P}} \mathcal{L}_{struct}^{k} +
\frac{1}{\lvert Q \rvert}
\sum_{k\in{Q}} \mathcal{L}_{text}^{k}
$$
$P$ is the set of layers which consider structure loss while $Q$ is for texture loss.
And for brevity, we use $(p, q)$ to represent the choice of $P,Q$,
which takes last $p$ layers as $P$ and last $q$ layers as $Q$.
For example, $(2, 1)$ represent $P = \{1, 2\}$ and $Q = \{1\}$,
which means completion results from last two layers of U-Net will be used to
compute structure loss and only last one for texture loss.
Corresponding part will be ignored if the total number of layers
$\lvert P \rvert$ or $\lvert Q \rvert$ equal to zero.
We will discuss the choice of $P$ and $Q$ in Section \ref{sec:analysis}.

\begin{figure*}[htb]
	\centering
	\begin{subfigure}[b]{\textwidth}
		\centering
	    \includegraphics[width=\textwidth]{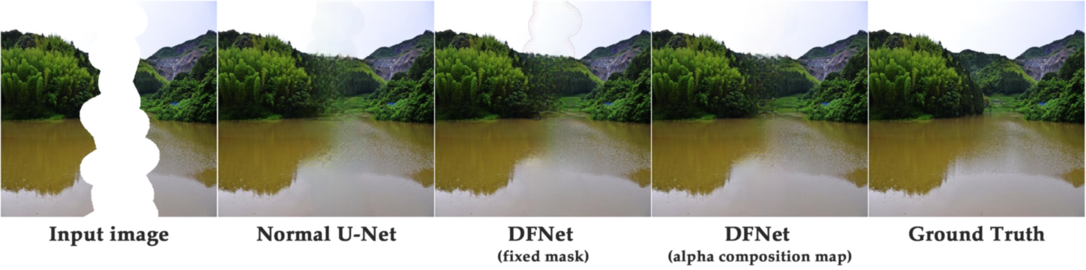}
    	\label{fig:effectiveness}
	\end{subfigure}
	\begin{subfigure}[b]{\textwidth}
		\centering
	    \includegraphics[width=\textwidth]{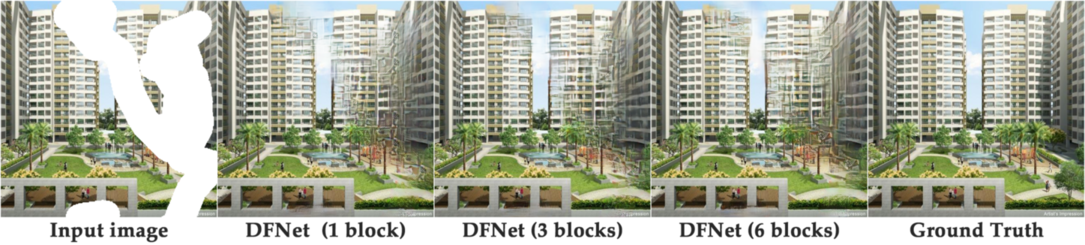}
	    \label{fig:multilayer}
	\end{subfigure}
	\begin{subfigure}[b]{\textwidth}
		\centering
	    \includegraphics[width=\textwidth]{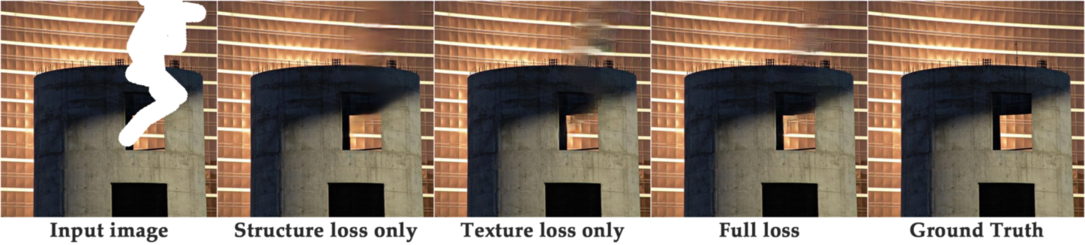}
	    \label{fig:losstune}
	\end{subfigure}
	\caption{Effectiveness of Fusion Block, Multi-scale constraints, and Loss Ablation. (a) compares the results of the proposed network without and with a fixed mask, and with fusion block. (b) depicts the results with 1, 3, 6 fusion blocks respectively. (c) shows the effects of structure loss and texture loss, and proves the effectiveness of combination loss.}
	\label{fig:exp}
\end{figure*}



\begin{figure*}
	\centering
	\includegraphics[width=\linewidth]{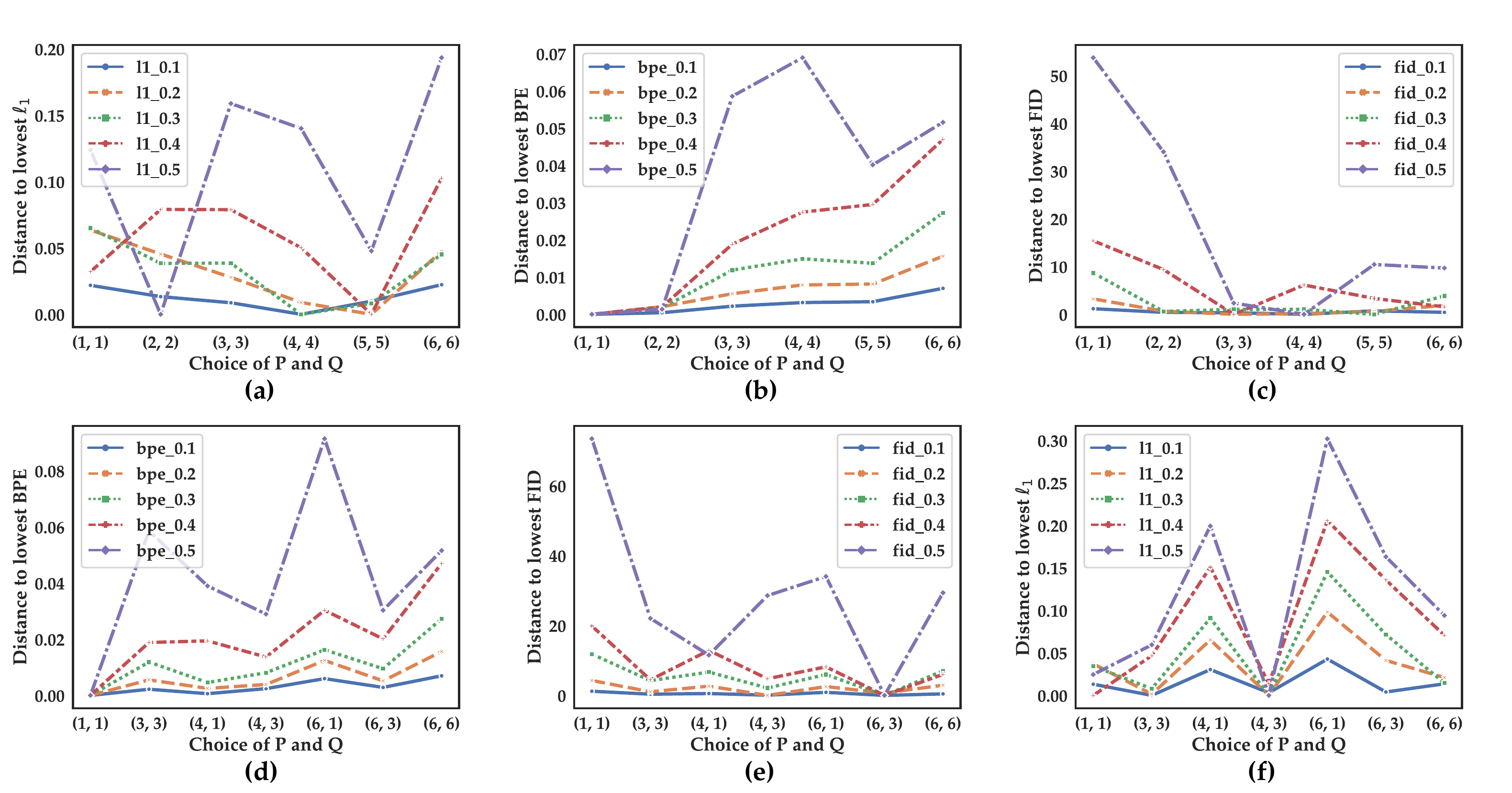}

	\caption{Evaluation of different $P$ and $Q$. Detailed description for $P$ and $Q$ can be found in Section \ref{sec:loss}.
	We separately compare models with different area range of mask.
	The results have been normalized by subtracting
	the minimal value of corresponding comparisons.
	For these three metrics, lower means better performance.
	In (a), (b) and (c), we compare 6 choices of $P$ and $Q$
	which gradually add fusion blocks but keeping $P = Q$.
	Results show performance gets better with multi-scale constraints.
	We further compare other 4 choices of $P,Q$ which $P \neq Q$ and
	choose $(6, 3)$ as our best model. More analysis can be found in Section \ref{sec:multiscale}
	and Section \ref{sec:loss_tune}.}
	\label{fig:choice}
\end{figure*}

\section{Experiments}

\subsection{Experiment Details}
We evaluate DFNet on two public datasets: \textbf{Places2}\cite{zhou2017places} and \textbf{CelebA} \cite{liu2015faceattributes}. For Places2, we use the original train, test, and val splits. For CelebA, we randomly partition into 27K images for training and 3Kimages for testing. Images in Places2 and CelebA are respectively resized to $512\times 512$ and $256 \times 256$ during training and testing. We randomly generate 1000 masks according to the method in \cite{yu2018free} and perform augmentation to these masks during training. To analysis the influence of unknown region range, these masks are categorized into five classes, including [0 10\%), [10\%, 20\%), [20\%, 30\%), [30\%, 40\%), [40\%, 50\%).

Models are separately trained on each dataset. Our proposed model is implemented in PyTorch\cite{paszke2017automatic} and trained in a single machine with 8 GeForce RTX 2080 Ti. We use Horovod\cite{alex2018horovod} as our distributed training framework. With a batch size of 6 for each GPU, it usually takes about 3 days to train a model. Forwarding is extremely fast, it only takes $8.29 ms$ to complete an image. As a common configuration, Adam\cite{kingma2014adam} is applied for optimization. The learning rate reduced from $2e-3$ to $2e-6$ in 20 epochs, with a decay rate of $0.1$ and step size of $5$.

\subsection{Evaluation Metrics}
Different from the tasks as image classification, detection and segmentation, image generation usually don't have strict targets. The basic rule is visually plausible. For image completion, it requires the completion not only looks real but also transit naturally from known region. So we apply $\ell_1$, and \textit{Fréchet Inception Distance} (FID) \cite{heusel2017gans} as evaluation metrics both in perspective of pixels and features to quantitatively analysis the performance of DFNet.

Furthermore, we observe that pixels in unknown region that near the boundary have very small variance while these pixels play the most important role in structure and texture transition. To measure the transition performance of models, we propose \textit{Boundary Pixels Error} (BPE) which only consider pixels error near the boundary. For boundary area $\mathbf{b}$, which is $n$ pixels narrow band adjacent to the boundary of unknown region, BPE is mean absolute error of those pixels between ground truth $\mathbf{I}$ and prediction $\hat{\mathbf{I}}$: $$
BPE = \frac{\lVert \mathbf{b} \odot (\mathbf{I} - \hat{\mathbf{I}}) \rVert_1}{\lVert \mathbf{b} \rVert_1}
$$

\begin{table*}
	\begin{center}
	\tabcolsep=0.128cm
		\begin{tabular}{lrrrrrrrrrrrrrrr}
			\toprule
			\multirow{2}{*}{\textbf{Methods}} & \multicolumn{5}{c}{\boldmath${\ell_1}$} & \multicolumn{5}{c}{\textbf{BPE}} & \multicolumn{5}{c}{\textbf{FID}} \\ \cmidrule(lr){2-6}  \cmidrule(lr){7-11}  \cmidrule(lr){12-16}
			{} &  0.1 &  0.2 &  0.3 &  0.4 &  0.5 &  0.1 &  0.2 &  0.3 &  0.4 &  0.5 &  0.1 &  0.2 &  0.3 &  0.4 &  0.5 \\
			\midrule
			\textbf{DeepFill}     &    2.79 &    6.75 &   10.63 &   15.35 &   28.38 &     1.33 &     1.81 &     2.39 &     2.91 &     5.13 &    24.04 &    56.55 &    98.25 &   173.90 &   324.97 \\
			\textbf{PConv}        &    1.51 &    4.22 &    7.01 &   10.52 &   12.83 &     0.17 &     0.37 &     0.62 &     0.87 &     1.60 &    14.98 &    41.21 &    84.60 &   166.72 &   217.48 \\
			\textbf{EdgeConnect}         &    1.43 &    3.94 &    \textbf{6.41} &    \textbf{9.64} &   \textbf{11.38} &     0.33 &     0.69 &     1.11 &     1.48 &     2.55 &    19.24 &    35.91 &    68.29 &   131.16 &   147.51 \\
			\textbf{DFNet}        &    \textbf{1.40} &    \textbf{3.91} &    6.50 &    9.89 &   11.96 &     \textbf{0.15} &     \textbf{0.33} &     \textbf{0.55} &     \textbf{0.74} &     \textbf{1.42} &    \textbf{12.27} &    \textbf{34.64} &    \textbf{65.25} &   \textbf{127.58} &   \textbf{136.22} \\
			\bottomrule
		\end{tabular}
	\caption{Quantitative comparison with other methods on Places2.}
	\label{tab:compare_other}
	\end{center}
\end{table*}

\subsection{Analysis of DFNet architecture} \label{sec:analysis}

In this section, we investigate the performance of the proposed modules in DFNet. First, we show the effectiveness of fusion blocks. Then we focus on the effect of multi-scale constraints by gradually increasing fusion blocks to DFNet and evaluating it. Finally, we discuss how to apply structure loss and texture loss on different resolution completion results to achieve the best results. 

\subsubsection{Effectiveness of Fusion Block.} \label{sec:effectiveness}
We compare our results with predictions from a normal U-Net and predictions from a DFNet but directly using mask to replace alpha composition map. We use only one fusion block for fair comparison, which means $P = Q = \{1\}$ for $\mathcal{L}_{total}$.(Section \ref{sec:loss}).

As can be seen in the 1st row of Figure \ref{fig:exp}, fusion block leads to the best transition near the boundary. Although most of semantic information has been restored, there exists obvious color transition inconsistent in the result of standard network without mask constraints. This is because global semantic consistency constraints can only leads to similar texture in the missing areas, but structural consistency can not be guaranteed. Based on the mask constraints, the pixel transition in filling area becomes more natural, which proves the effect of the proposed method on the propagation of structural and texture information. As mentioned above, the alpha composition map is a attention mechanism to enhance the structural consistency. Furthermore, the result of learned alpha mapping is even better in the edge transition to eliminate the visible artifacts near the boundary. 

The same detailed conclusion can be seen in Figure \ref{fig:sample-alpha}. Based on the proposed fusion block, the structure between the known and unknown areas are well preserved, even beyond the mask area. The sharp edge of the roof is retained into the reconstructed image with other useless part discarded.

\subsubsection{Multi-scale constraints.} \label{sec:multiscale}

We compare DFNets with different number of fusion blocks from one to six.
Formally speaking, $P$ and $Q$ in Section \ref{sec:loss} both increase from $\{1\}$ to $\{6\}$.
In this section, $P$ and $Q$ are equal to only analyze the role of multi-scale fusion. 

As can be seen in the 2nd row of Figure \ref{fig:exp}, the structure of building is more clear and accurate based on more fusion blocks. Also the shapes of houses are depicted in the result of 3 fusion blocks instead of the noises in the result based on 1 fusion block. While high level layers in encoder have bigger receptive field and global context, the structure information can be more easily reconstructed with more layers in decoder.
Nevertheless, although the result of 6 fusion blocks retains these structural information, its texture is not very stable compare to 3 fusion blocks. We guess this is because we shouldn't apply texture constraints for low resolution completion result. In the next section, we will go into more detail about how to choose the number of blocks layers. 

We also give the quantitatively comparison in the second row of Figure \ref{fig:choice}. Results are separated according to the range of mask in each evaluation metric. With fusion blocks increased, FID gets lower and lower. This means multi-layer constraints helps to capture contextual information and makes the whole image looks more real. The BPE increases slightly with fusion blocks increased. This can be explained that finer texture and smoother transition is a trade-off. However, globally visual effect is more important and the change in BPE actually is very small.

\subsubsection{Loss Ablation and Tuning.} \label{sec:loss_tune}

Firstly, the effect of structure loss $\mathcal{L}_{struct}$ and texture loss are showed  $\mathcal{L}_{text}$ by respectively trained DFNet only applies only one of them. As seen in the 3rd Figure \ref{fig:exp}, the result without texture loss is blurry but with accurately structure consistency, while the other one completely destroys the structure, it fails to recover edges of object although they have finer textures. This provides strong evidence for loss design in this paper.

We further discuss the dynamic loss design in each layer. Based on the visualization results in \ref{sec:multiscale}, we make a comprehensive comparison of the loss design in different layers. As shown in Figure \ref{fig:choice}, the performance depicts the same trend with different ranges of hole size. We choose $P = \{1, 2, 3, 4, 5, 6\}$ to compute structure loss and $Q = \{1, 2, 3\}$ for texture loss in the final architecture. This can be explained that, although the structure information is more and more abundant with higher and higher encoder layers, the high-level features will lead to texture noise due to the loss of global semantic constraints.

\begin{figure*}
	\captionsetup[subfigure]{labelformat=empty}
	\centering
	\begin{subfigure}[b]{0.97\textwidth}
		\centering
		\includegraphics[width=\textwidth]{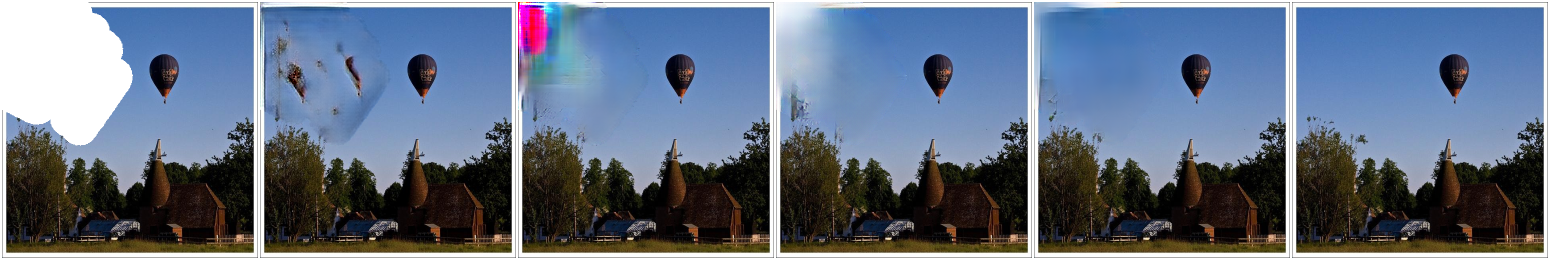}
	\end{subfigure}
	\begin{subfigure}[b]{0.97\textwidth}
		\centering
		\includegraphics[width=\textwidth]{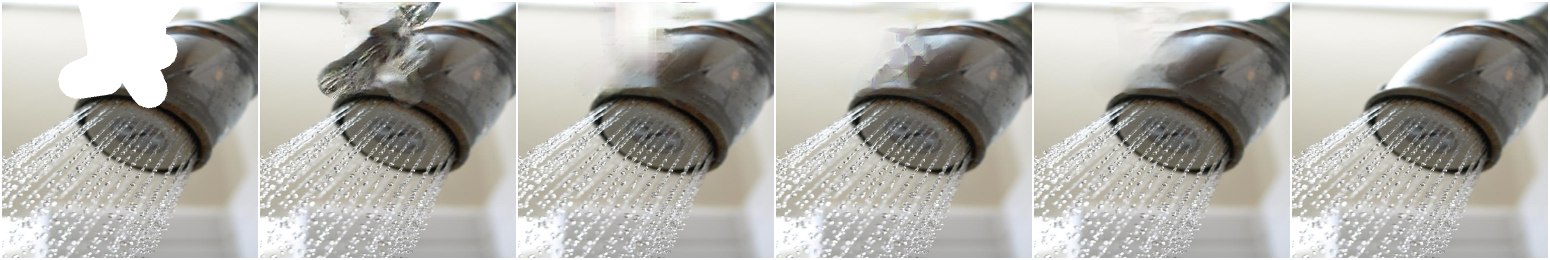}
	\end{subfigure}
	\begin{subfigure}[b]{0.97\textwidth}
		\centering
		\includegraphics[width=\textwidth]{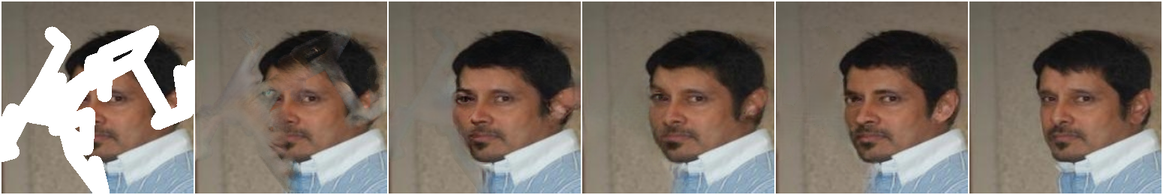}
	\end{subfigure}
	
	\begin{subfigure}[b]{0.16\textwidth}
		\centering
		\caption{Input image}
	\end{subfigure}
	\begin{subfigure}[b]{0.16\textwidth}
		\centering
		\caption{DeepFill\cite{yu2018generative}}
	\end{subfigure}
	\begin{subfigure}[b]{0.16\textwidth}
		\centering
		\caption{PConv\cite{liu2018partialinpainting}}
	\end{subfigure}
	\begin{subfigure}[b]{0.16\textwidth}
		\centering
		\caption{Edge Connect\cite{nazeri2019edgeconnect}}
	\end{subfigure}
	\begin{subfigure}[b]{0.16\textwidth}
		\centering
		\caption{Ours DFNet}
	\end{subfigure}
	\begin{subfigure}[b]{0.16\textwidth}
		\centering
		\caption{Ground Truth}
	\end{subfigure}
	
	\caption{comparison results on Places2 and CelebA. More results can be found in the supplementary materials.
	}
	\label{fig:compare_places2}
\end{figure*}

\subsection{Comparisons with Other Methods}

We quantitatively and qualitatively compare our DFNet with 3 recently methods,
including DeepFill \cite{yu2018generative}, PConv \cite{liu2018partialinpainting} and Edge Connect\cite{nazeri2019edgeconnect}.
Results of DeepFill and Edge Connect are obtained by using their pre-trained models \footnote{\href{https://github.com/JiahuiYu/generative\_inpainting}{https://github.com/JiahuiYu/generative\_inpainting}} \footnote{\href{https://github.com/knazeri/edge-connect/}{https://github.com/knazeri/edge-connect/}}. However, we don't find the official implementation of PConv,
so we implement one with the same settings described in the original paper.

\subsubsection{Quantitative Comparisons.}
Table \ref{tab:compare_other} shows the comparison results on Places2\cite{zhou2017places}. We use three metrics including $\ell_1$,
BPE and \textit{Fréchet Inception Distance} (FID) \cite{heusel2017gans}.
Results from ours outperforms others on both boundary transition and realistic on overall image.
Our predictions on BPE is significantly lower than those
from Edge Connect\cite{nazeri2019edgeconnect} and other methods.
This means completion from our methods have better transitional area near the boundary, which also proves the effectiveness of proposed fusion blocks.

Edge Connect works well on maintaining structural consistency
by applying additional edge constraints.
However it doesn't pay much attention to smooth transition.
The constraints on the structure of the whole image
can't lead to natural image restoration, especially in detail.
Results of Edge connect shows lower $\ell_1$ than ours while the missing hole is large.
But this only state results of Edge Connect is more similar to original images.
Because completion can be more diverse while hole is larger.

PConv use partial convolution to progressively reduce missing region,
which can be considered as providing a hidden attention map
gradually enlarged from boundary area to full known region.
This enhance the learning ability near the boundary,
which have the similar effects with the proposed DFNet when considering transition performance.
However, this architecture is not good at large hole
because information can't be transmitted effectively to inner area.
When comparing PConv and Edge Connect on BPE and FID,
we can find PConv has better transition near the boundary than Edge Connect
and comparable FID when missing hole is small,
however, when missing hole becomes larger, Edge Connect will have more realistic results.

\subsubsection{Qualitative Comparisons.}

Figure \ref{fig:compare_places2} shows the comparison on Places2 and CelebA without any post-processing. As shown in the figure, we can see our model has the best performance in texture consistency near the boundary, and also good at keeping the structure consistency even better than Edge Connect. Results from different datasets shows the generalization ability of our methods.

There is one thing should be noticed, as shown in the 1st case of Figure \ref{fig:compare_places2},
we find PConv and Edge Connect sometimes fail to complete the missing hole when the missing hole cover the border of an image. For PConv, we think this is the limit of partial convolution,
which can't transmit information into a very large hole. While for Edge Connect, it always produces clouds like completion in similar situation. We couldn't figure it out the reason.

\section{Conclusion}
In this paper, we analysis the image completion technology from a new perspective. We propose Deep Fusion Network by designing a fusion block to predict an alpha composition map for combining completion and existing content and embedding it on multi-scale layers. Results of experiments on Places2 and CelebA dataset shows our method achieves state-of-the-art results, especially in the filed of harmonious texture transition, texture detail and semantic structural consistency. 

{\small
\bibliographystyle{ieee_fullname}
\bibliography{main}
}

\appendix


\begin{figure*}[p]
	\captionsetup[subfigure]{labelformat=empty}
	\begin{subfigure}[b]{0.97\textwidth}
		\centering
		\includegraphics[width=\textwidth]{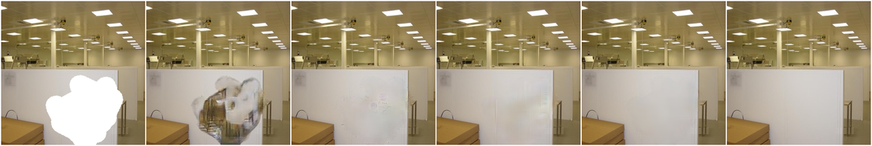}
	\end{subfigure}
	\begin{subfigure}[b]{0.97\textwidth}
		\centering
		\includegraphics[width=\textwidth]{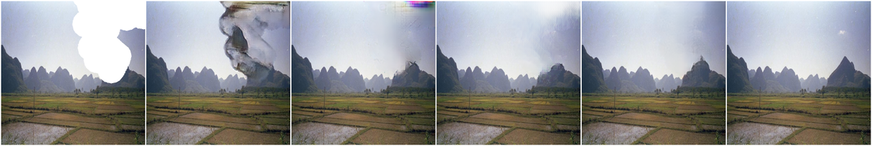}
	\end{subfigure}
	\begin{subfigure}[b]{0.97\textwidth}
		\centering
		\includegraphics[width=\textwidth]{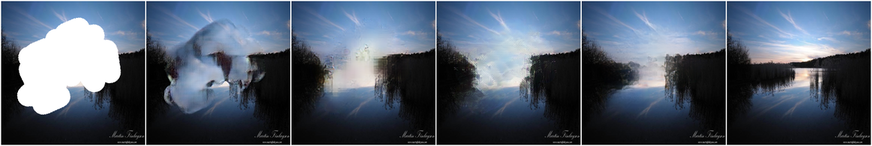}
	\end{subfigure}
	\begin{subfigure}[b]{0.97\textwidth}
		\centering
		\includegraphics[width=\textwidth]{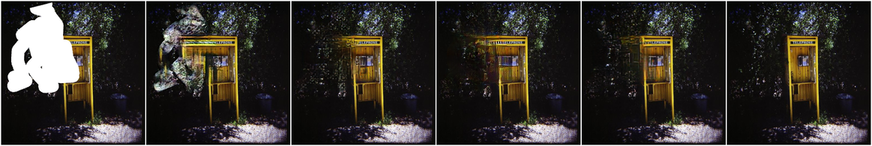}
	\end{subfigure}
	\begin{subfigure}[b]{0.97\textwidth}
		\centering
		\includegraphics[width=\textwidth]{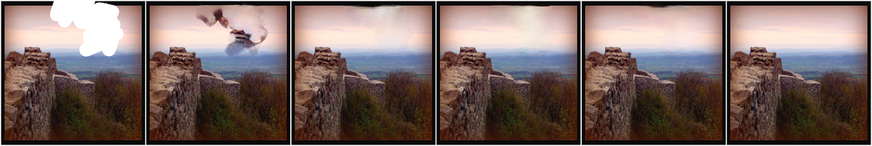}
	\end{subfigure}
	\begin{subfigure}[b]{0.97\textwidth}
		\centering
		\includegraphics[width=\textwidth]{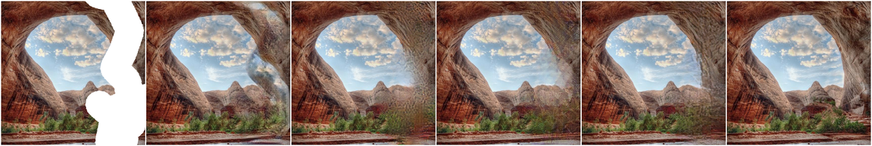}
	\end{subfigure}
	\begin{subfigure}[b]{0.97\textwidth}
		\centering
		\includegraphics[width=\textwidth]{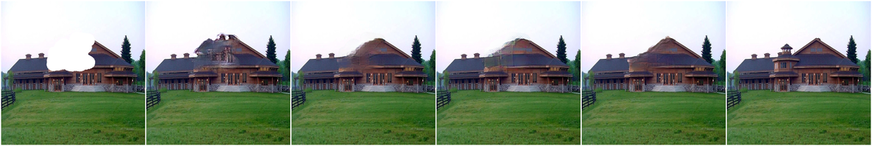}
	\end{subfigure}
	
	\begin{subfigure}[b]{0.16\textwidth}
		\centering
		\caption{Input image}
	\end{subfigure}
	\begin{subfigure}[b]{0.16\textwidth}
		\centering
		\caption{DeepFill}
	\end{subfigure}
	\begin{subfigure}[b]{0.16\textwidth}
		\centering
		\caption{PConv}
	\end{subfigure}
	\begin{subfigure}[b]{0.16\textwidth}
		\centering
		\caption{Edge Connect}
	\end{subfigure}
	\begin{subfigure}[b]{0.16\textwidth}
		\centering
		\caption{Ours DFNet}
	\end{subfigure}
	\begin{subfigure}[b]{0.16\textwidth}
		\centering
		\caption{Ground Truth}
	\end{subfigure}
	
	\caption{comparison results on Places2.}
	\label{fig:compare_places2}
\end{figure*}

\begin{figure*}[p]
	\captionsetup[subfigure]{labelformat=empty}
	\centering
	\begin{subfigure}[b]{0.97\textwidth}
		\centering
		\includegraphics[width=\textwidth]{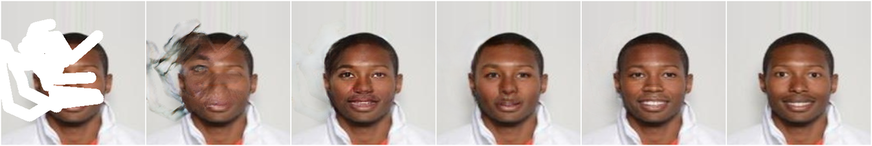}
	\end{subfigure}
	\begin{subfigure}[b]{0.97\textwidth}
		\centering
		\includegraphics[width=\textwidth]{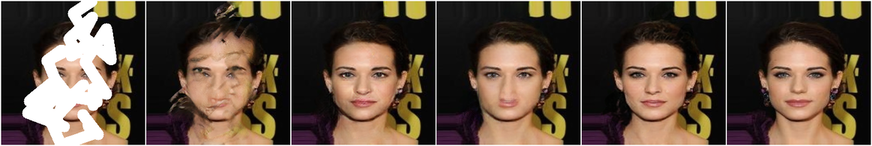}
	\end{subfigure}
	\begin{subfigure}[b]{0.97\textwidth}
		\centering
		\includegraphics[width=\textwidth]{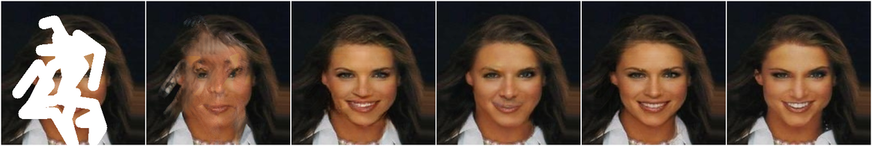}
	\end{subfigure}
	\begin{subfigure}[b]{0.97\textwidth}
		\centering
		\includegraphics[width=\textwidth]{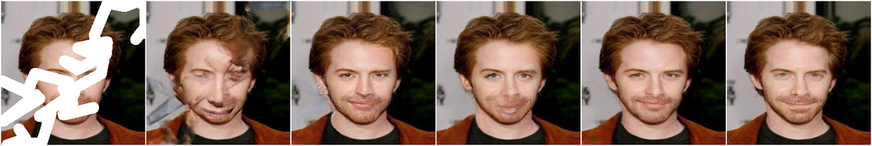}
	\end{subfigure}
	\begin{subfigure}[b]{0.97\textwidth}
		\centering
		\includegraphics[width=\textwidth]{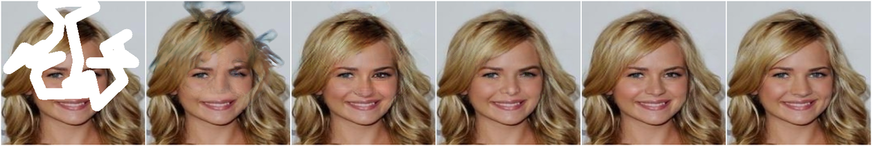}
	\end{subfigure}
	\begin{subfigure}[b]{0.97\textwidth}
		\centering
		\includegraphics[width=\textwidth]{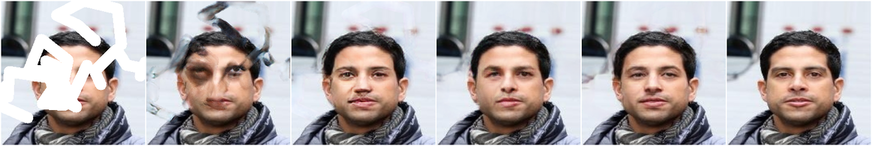}
	\end{subfigure}
	\begin{subfigure}[b]{0.97\textwidth}
		\centering
		\includegraphics[width=\textwidth]{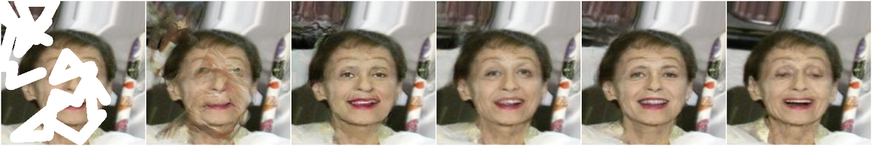}
	\end{subfigure}
	
	\begin{subfigure}[b]{0.16\textwidth}
		\centering
		\caption{Input image}
	\end{subfigure}
	\begin{subfigure}[b]{0.16\textwidth}
		\centering
		\caption{DeepFill}
	\end{subfigure}
	\begin{subfigure}[b]{0.16\textwidth}
		\centering
		\caption{PConv}
	\end{subfigure}
	\begin{subfigure}[b]{0.16\textwidth}
		\centering
		\caption{Edge Connect}
	\end{subfigure}
	\begin{subfigure}[b]{0.16\textwidth}
		\centering
		\caption{Ours DFNet}
	\end{subfigure}
	\begin{subfigure}[b]{0.16\textwidth}
		\centering
		\caption{Ground Truth}
	\end{subfigure}
	
	\caption{comparison results on CelebA.
	}
	\label{fig:compare_places2}
\end{figure*}

\end{document}